\documentclass[preprint]{article}

\usepackage[nonatbib]{neurips_2024}

\usepackage[square,numbers,sort]{natbib}
\usepackage[utf8]{inputenc} % allow utf-8 input
\usepackage[T1]{fontenc}    % use 8-bit T1 fonts
\usepackage{hyperref}       % hyperlinks
\usepackage{url}            % simple URL typesetting
\usepackage{booktabs}       % professional-quality tables
\usepackage{amsfonts}       % blackboard math symbols
\usepackage{nicefrac}       % compact symbols for 1/2, etc.
\usepackage{microtype}      % microtypography
\usepackage{xcolor}         % colors
\usepackage{amsmath}
\usepackage{algorithm}
\usepackage{algpseudocode}

\usepackage{graphicx} 
\usepackage{wrapfig}
\usepackage{array} % 提供更多的列格式化选项

\usepackage{makecell}
\usepackage{multirow}
\usepackage{tabularx}
\usepackage{array}

\definecolor{darkblue}{rgb}{0.12, 0.56, 1.00}

\newcommand{\oursystem}{Grade-Like-a-Human}
\newcommand{\ourdataset}{OS}

\title{Grade Like a Human: Rethinking Automated Assessment with Large Language Models}

\author{
  Wenjing Xie  \thanks{OS dataset available at: \url{https://github.com/wenjing1170/llm_grader}}\\
  City University of Hong Kong \\
  wenjing.xie@my.cityu.edu.hk \\
  \And
  Juxin Niu \\
  City University of Hong Kong \\
  juxinniu@cityu.edu.hk \\
  \And
  Chun Jason Xue \\
  Mohamed bin Zayed University of Artificial Intelligence \\
  jason.xue@mbzuai.ac.ae\\
  \And
  Nan Guan\\
  City University of Hong Kong\\
  nanguan@cityu.edu.hk
}

\begin{document}
\maketitle

\begin{abstract}
\label{s:abstract}

\iffalse
虽然LLMs已经被应用到自动grading system中，但是在处理复杂问题上还没有达到人类水平。
现有的工作都是在给定事先制定好的grading rubric的前提下去研究如何打分。
但是grading rubric本身的质量也会很大程度的影响grading质量，并且不管是LLMs还是人类打分者都不能完全保证自己提供的grading结果是完全正确且公平的。因此，打分任务是一个系统的复杂任务，目前还没有相关工作对打分任务进行系统性研究。
为了保证打分结果的的正确性，一致行和公平性，应该从以下三个方面进行考虑：
1）制定既能体现问题考察要点， 又考虑实际学生回答情况的grading rubric。这样的rubric才能消除制定好的grading rubric与实际学生回复的偏差，指导出来的打分结果才能很好的评估学生的实际表现。
2）在grading rubric指导下，打出准确且一致的分数，并且对每个学生要提供customized feedback。
3）在打分阶段完成后，要对打分结果进行复查。
我们论文首次提出了一个打分系统，包含以上三种能力。
此外，我们在实际的operating system课程上收集了一个datast，called "OS" . 并且在OS dataset上和广泛使用的Mohler dataset上开展了大量实验，来证明我们提出的效果的有效性, 为基于LLMs开发自动打分系统提供了一种新的，系统的insights.
\fi

While large language models (LLMs) have been used for automated grading, they have not yet achieved the same level of performance as humans, especially when it comes to grading complex questions. 
Existing research on this topic focuses on a particular step in the grading procedure: grading using predefined rubrics. However, grading is a multifaceted procedure that encompasses other crucial steps, such as grading rubrics design and post-grading review. There has been a lack of systematic research exploring the potential of LLMs to enhance the entire grading~process.

In this paper, we propose an LLM-based grading system that addresses the entire grading procedure, including the following key components:
1) Developing grading rubrics that not only consider the questions but also the student answers, which can more accurately reflect students' performance.
2) Under the guidance of grading rubrics, providing accurate and consistent scores for each student, along with customized feedback.
3) Conducting post-grading review to better ensure accuracy and fairness.
Additionally, we collected a new dataset named \ourdataset~from a university operating system course and conducted extensive experiments on both our new dataset and the widely used Mohler dataset. Experiments demonstrate the effectiveness of our proposed approach, providing some new insights for developing automated grading systems based on LLMs.

% Although LLMs have been applied in automated grading systems, they have not yet achieved human-level performance in handling complex questions. 
% Existing works on this topic mainly focused on apply prompt engineering to improve the grading performance but lacks a comprehensive research of the entire assessment task.

% In this paper, we explore the potential of utilizing LLMs to grade systematically in a human-judge manner.
% Effective evaluation of question answers needs to guarantee correctness, consistency, and fairness. To achieve this goal, we propose that an ideal assessment system using LLMs should behave like expert human judges with three key abilities necessary: generating grading criteria, grading with reference, and reviewing scores.
% We collect a dataset called ``OS'' from a real operating system teaching course and conduct extensive experiments using both the OS dataset and the publicly available Mohler dataset. Our experiments demonstrate that LLMs possess remarkable capabilities in addressing various subtasks within the assessment process, offering valuable insights for the development of a human-like assessment system. 

\end{abstract}

\section{Introduction}
\label{s:intro}

Large Language Models (LLMs)~\citep{llm-glm,llm-gpt3,llm-llama2,llm-mistral,llm-opt} have significantly revolutionized the world academically and in industry by excelling in multiple Natural Language Processing (NLP) tasks~\citep{hugginggpt,metagpt,toolllm,gorilla}. 
These models possess a profound understanding of a vast array of knowledge, can perform efficient logical reasoning, and exhibit deep comprehension of natural language. 
Therefore, they have been extensively integrated into workflows across numerous sectors worldwide.

A promising application in higher education involves leveraging LLMs to enhance the \textit{grading} process for students~\citep{fagbohun2024beyond, divya2023automation, chang2024automatic}. 
The grading process involves assessing their work and performance, providing critical feedback, and finally assigning scores. 
As a key measure of evaluating students' learning progress, scores must be given accurately, consistently, and fairly.
Traditionally, grading requires educators to design a comprehensive rubric and meticulously review each student's answers, a process that is both time-consuming and labor-intensive~\citep{jauhiainen2024evaluating}. As a result, there is growing interest in using LLMs to assist in this process.

Using LLMs to assist the grading process can be categorized under the research domain of Automated Short Answer Grading (ASAG) systems~\citep{fagbohun2024beyond, divya2023automation,chang2024automatic,yoon2023short}.
Current state-of-the-art systems typically grade each answer according to predefined grading rubrics set as system instructions and provide potential feedback. However, they still face significant challenges in achieving optimal grading performance in the following aspects:
\begin{itemize}
    \item \textit{Rubric Generation}~\citep{wang2024enhancing}: 
    Existing systems require educators to meticulously craft detailed grading rubrics for each question in advance. 
    This process requires a significant investment of effort and time, and it is further complicated by the fact that even minor variations in grading rubrics can result in markedly different grading results~\citep{doostmohammadi2024reliable}; and educators cannot predict which rubric will be the most effective.
    Additionally, the design of rubrics and the grading process are currently conducted independently. This separation means that rubrics cannot be tailored to consider students' actual answers during the design phase. For instance, in the case of open-ended questions, students' responses may not fall within the scope of the rubric's guidelines, making the grading results unreliable.
    \item \textit{Consistency \& Fairness}~\cite{achintalwar2024detectors,raina2024llm,liu2024aligning}: 
    An answer should receive the same score across multiple independent grading sessions; and similar answers should receive similar scores. 
    However, due to the inherent randomness and hallucinations~\cite{huang2023survey} of LLMs, the model might produce entirely different scores for the similar input. 
    Besides, existing ASAG systems grade each answer independently, lacking the means to evaluate the overall fairness of the grading process and to further optimize it.
\end{itemize}
The complexity and open-ended nature of the questions being graded exacerbate these issues~\citep{cohn2024chain,jauhiainen2024evaluating}. For instance, some open-ended questions do not have a standard answer that can be well-defined with a single statement; complex questions may require multiple sub-steps for grading; and lengthy questions and answers might exceed the LLM's context length.
These challenges motivate us to propose a better approach to grading.

Human excellence in grading stems from their ability to perform \textit{systematic} assessments, which involve phased planning, adjustments, and reflection throughout the grading process. 
Specifically: 
1) Humans can not only create standard rubrics based on the questions but also refine these rubrics by considering students' responses, thereby enhancing their comprehensiveness.
2) During grading, humans can accurately apply these rubrics to provide accurate scores. They can also remember previous student scores, which aids in maintaining fairness and consistency.
3) After completing grading, humans can review the results, identify any unreasonable outcomes through comparison and reflection, and make necessary adjustments.

\begin{figure}[]
  \centering
  \includegraphics[width=1.\textwidth]{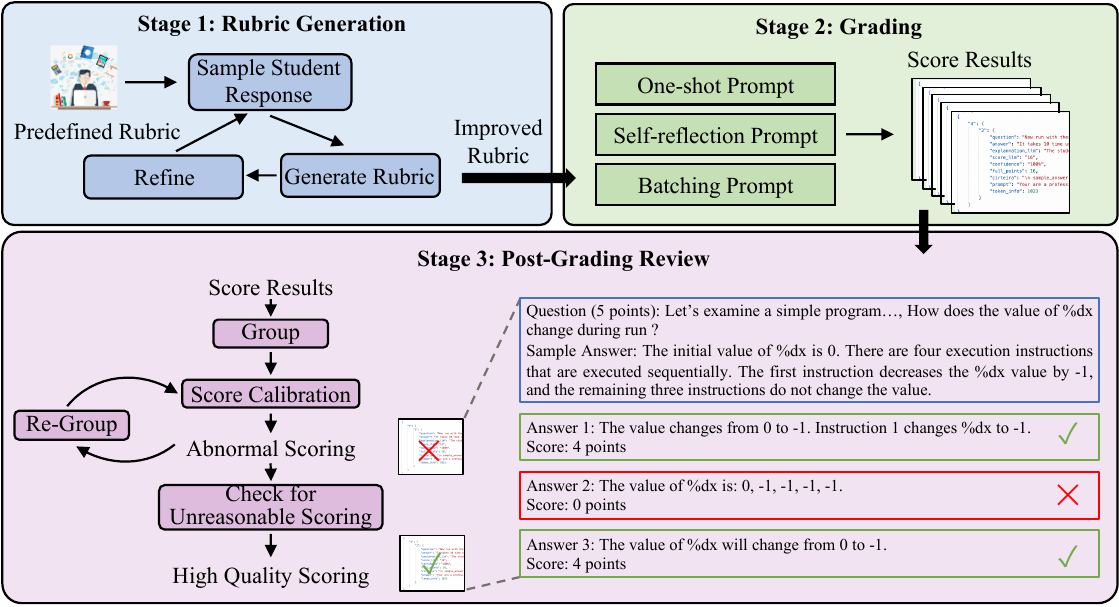} 
  \caption{Overview of the {\oursystem}~system}
  \label{fig:sys_framework}
\end{figure}

Inspired by human grading methodologies, \textit{we propose a multi-agent grading system named \textbf{\oursystem} to rethink the grading task in a systematic manner}. 
Specifically, the grading process is divided into three stages: \textit{rubric generation}, \textit{grading}, and \textit{post-grading review}.
In the rubric generation stage, we incorporate students' answers into the rubric design. By introducing a sampling-based iterative generation method, we optimize the generation process to make rubrics more targeted and effective. 
The optimized rubrics are then used in the grading stage to guide LLMs in grading students' answers. During this process, we further explore the impact of various prompt strategies on performance.
Finally, in the post-grading review stage, all students' grading results are reviewed. Through a group comparison approach, the LLM identifies unreasonable results, which can then be sent back for re-grading. Figure \ref{fig:sys_framework} illustrates the framework of the system.

Furthermore, \textit{we have collected and open-sourced a dataset named \textbf{\ourdataset} for evaluating the performance of LLMs in grading tasks}. 
This dataset is derived from our undergraduate Operating Systems course and includes all questions from the tutorials and assignments, along with the students' answers. Each answer is also labeled with human-given scores.
We utilized this dataset, along with an open-source dataset Mohler~\cite{mohler2011learning}, to thoroughly evaluate our system. The experimental results demonstrated our system's excellent capability in grading tasks.

Our contributions are summarized as follows: 
We point out that current ASAG systems lack a systematic perspective in viewing the grading task; and therefore face challenges in rubric generation, grading consistency, and fairness for complex questions.
To address these issues, we are the first to propose systematically designing the assessment process to explore the potential of LLMs; and further introduce a multi-agent grading framework named \oursystem, which has the ability to plan, reflect, and adjust in multiple stages in grading tasks. To effectively evaluate the system's performance, we collected and open-sourced the \ourdataset~dataset. Comprehensive experiments conducted on this dataset and the Mohlar dataset demonstrate that our system can significantly improve the accuracy and reliability of grading tasks.

\section{Related Work}

There has been a surge in research interest in the field of LLMs due to their exceptional performance across a wide range of tasks~\citep{brown2020language, thoppilan2022lamda, chowdhery2023palm, hoffmann2022training, OpenAI_report}. A crucial application is utilizing LLMs as graders for automatic answer grading~\citep{fagbohun2024beyond, divya2023automation, chang2024automatic}.
Existing works~\citep{pinto2023large, song2024automated, henkel2023can} have demonstrated the capability of ChatGPT in understanding semantic details to grade open-ended question answers and provide feedback. For instance, \citet{nilsson2023gpt} used GPT-4 to grade programming assignments, proving the effective grading capabilities of LLMs for code-based responses.
Extensive research efforts have also focused on leveraging advanced prompt engineering to enhance the grading results of LLMs. \citet{yoon2023short} employed one-shot prompts to provide both analytic scores and final holistic scores for short answers. \citet{cohn2024chain} proposed Chain-of-Thought prompts to enable LLMs to reason and evaluate the steps of student scientific experiments. Comparative prompt~\citep{liusie2024llm} and Rankprompt~\citep{hu2024rankprompt} transform direct score grading tasks into comparative assessment tasks, using relative comparison prompts to leverage LLMs for grading. 
\citet{madaan2024self} introduced Self-Refine prompts to improve initial outputs from LLMs through iterative feedback and refinement. \citet{hasanbeig2023allure} presented ALLURE, an iterative approach that compares LLM-generated evaluations with annotated data and then incorporates human feedback for the next iteration. However, these iterative methods require annotated data and human feedback, which can be challenging to apply in real-world, large-scale answer grading scenarios. \citet{del2023gradeaid} proposed the GradeAid framework for real-time automatic short answer grading in educational contexts.
Additionally, some studies~\citep{latif2024fine, song2024automated} have explored fine-tuning LLMs for specific types of short answer questions, but this approach requires collecting large-scale datasets and fine-tuning LLMs, which can be a non-trivial task.

While LLMs have shown promising results in automatic grading tasks, there are still limitations in their real-world grading applications, and the quality of LLMs grading has not yet reached the level of professional human graders~\citep{ChallengeLLMJudge, chen2024humans}. \citet{liu2024aligning} studied the misalignment between LLM evaluators and human judgment, revealing that existing biases. \citet{doostmohammadi2024reliable} found that while automatic evaluation methods can approximate human judgments under specific conditions, their reliability is highly context-dependent. The papers~\citep{achintalwar2024detectors} and~\citep{raina2024llm} delve into the challenges of applying these models from the perspective of safety and reliability.

Distinct from these approaches, our proposed \oursystem~system is the first to conduct a systematic study of the grading process, which we have divided into three stages: grading rubric generation, grading, and post-grading review. LLMs collaborate as multi-agents across these three stages, ultimately producing accurate, consistent, and fair grading results.

% Existing methods predominantly adopt a single-agent approach to grading, without systematically decomposing complex assessment tasks. Distinct from these approach, we advocate that LLMs should grade like humans by exploring LLMs in assessment tasks from a systematic perspective and leveraging a multi-agent approach to coordinate multiple LLMs to enhance grading effectiveness.

% \iffalse
% 与这些方法不同的是，我们提出的Grade-like-a-human system是第一个对打分过程进行系统性研究，并将其分成grading rubric的制作，打分，以及分数复查三个阶段。基于LLMs以multi-agent的方式在三个阶段之间相互协作，最后输出正确，一致且公平的打分结果。
% \fi

% %%%%%%%%%% cite in introduction
% (end of related work)
% multi-agent in other application
% \cite{chan2023chateval}
% \cite{hong2023metagpt}
% \cite{seff2023motionlm}
% \cite{yang2023spatio}

% dataset

% mohler dataset \cite{mohler2009text, mohler2011learning}
% \cite{sonkar2024automated}
% \cite{zhuang2024toolqa}

\section{\oursystem}

\oursystem~is a multi-agent system designed to enhance the performance of LLMs in grading tasks. 
The system framework is illustrated in Figure~\ref{fig:sys_framework}.
The system consists of three stages: \textit{Rubric Generation}, \textit{Grading}, and \textit{Post-Grading Review}.
In the Rubric Generation stage, we refine the pre-established scoring criteria to enhance their accuracy and comprehensiveness. The optimized scoring criteria will be used for grading in the Grading stage. After grading is complete, in the Post-Grading Review stage, we examine all the grading results to ensure there are no unreasonable outcomes.
The subsequent sections will provide detailed explanations of each stage.

\subsection{Rubric Generation} 
\label{s-gencri}

In this section, we present our method for Rubric Generation.
The Rubric Generation stage takes place before the grading tasks begin.
Our aim is to optimize the rubric specifically based on students' actual answers, making the subsequent grading process more accurate and reliable.

The rubric generation process is completed through multiple iterations following a self-reflective paradigm~\cite{madaan2024self}.
Specifically, for each question to be graded, we have a pre-defined rubric created by humans and all the students' answers.
In each iteration, we will select a small sample from the students' answers and have humans score these answers and provide reasons for the scores. These scores and reasons will then be provided as samples to the LLMs to learn how to grade and to help formulate the rubric with improved qualities. This optimized rubric is then carried forward to the next iteration as the input. After the final iteration, the refined rubric is adopted as the final version and progresses to the subsequent stage.
Formally speaking, consider a question $q$, a pre-established rubric $r_0$, and a set of student answers $A = \{a_1, a_2, \ldots, a_n\}$. 
We define a sampling method $S$ such that $S: A \to A_m$, where $A_m \subseteq A$ and $|A_m| = m$ with $m \ll n$. After human grading, we obtain a set of answer-score pairs $\{(a_i, g_i) \mid a_i \in A_m\}$. 
Based on this, an iteration can be formalized as a process represented by 
\begin{equation}\label{alg:rubric-gen}
    r_{i+1} = O\left(r_i, \{(a_i, g_i)\}, p\right)
\end{equation}
where $p$ is the system prompt sent to the LLMs. 
By performing this process $t$ times, we derive an optimization sequence $\{r_0 \to r_1 \to \ldots \to r_t\}$, with $r_t$ being the final refined rubric.
The process $O$ involves combing $r_i$, $\{(a_i, g_i)\}$, and $q$ to form a prompt for the LLM to generate a refined rubric. We give the details in Appendix~\ref{appedix:prompt_generate_rurbric}.

% The process $O$ involves combining $r_i$, $\{(a_i, g_i)\}$, and $p$ to form the prompt. We give the details in Appendix~\ref{appedix:prompt_generate_rurbric}.
% \iffalse
% 过程O涉及将$r_i$, $\{(a_i, g_i)\}$, and $q$结合形成prompt，让LLM来生成refined rubric的。
% \fi

\paragraph{Sampling Methods}
The sampling method $S$ directly affects the quality of the rubric.
\textit{Random sampling} is the simplest method. In each iteration, we randomly select from the remaining students' answers. However, the sample points obtained through this method do not accurately reflect the distribution of student performance under the given grading rubric. 
We aim for the sampling results to adhere to this distribution, as such a sample distribution can better guide LLMs in learning the impact of the rubric on different types of student answers. Based on this, we propose the \textit{Score-Distribution-Aware Sampling} method.
The Score-Distribution-Aware Sampling method consists of two steps. 
In the first step, the LLM performs an initial grading of all student answers based on the current grading rubric and obtains a score for each answer. This initial results allow us to obtain the distribution of student scores under the current grading rubric, which can guide our sampling process.
In the second step, we conduct scoring based on this sampling. The goal is to ensure that the distribution of the $m$ selected samples closely matches the score distribution obtained from the initial assessment.
We use a stratified sampling algorithm~\cite{liberty2016stratified} to achieve this goal.
Formally, grading all student answers in set $A$ using the current rubric $r_i$ yields scores $\{g_j \mid a_j \in A\}$.
To sample $m$ points from $n$ total answers, we first divide these scores into $k$ strata $\{B_1, B_2, \ldots, B_k\}$, with each stratum $B_l$ containing answers within a specific score range. 
For each stratum $B_l$, we compute the proportion $p_l = |B_l|/n$. 
Then, we determine the number of samples for each stratum as $m_l = \lceil p_l \times m \rceil$. 
Finally, we randomly sample $m_l$ answers from each $B_l$ to form the sample set $A_m$ as follows.
\begin{equation}
    A_m = \bigcup_{l=1}^k \text{RandomSample}(B_l, m_l)
\end{equation}
In Appendix~\ref{appedix:sampling}, we provide the pseudocode implementations of these two sampling methods respectively.

\subsection{Grading}
\label{s-prompt}

In this section, we give the design of the Grading stage.
The grading stage adheres to the existing ASAG paradigm~\cite{divya2023automation}. 
For the evaluation process, the student's answers, the question description, the improved rubric from the previous stage, and any additional necessary context and instructions are compiled into a prompt. This prompt is then inputted into the LLM for grading.

\begin{figure}[]
    \centering
    \begin{minipage}[b]{0.327\textwidth}
        \includegraphics[width=\textwidth]{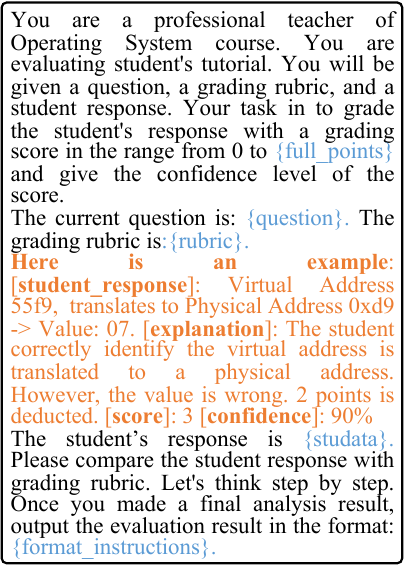}
        \caption{One-shot prompt} % 图1的标题
        \label{fig:oneshot}
    \end{minipage}
    \hfill
    \begin{minipage}[b]{0.327\textwidth}
        \includegraphics[width=\textwidth]{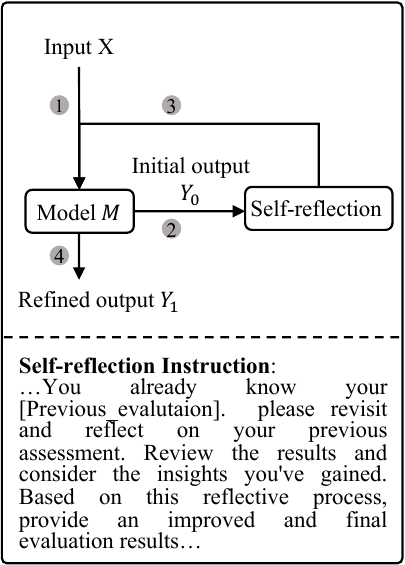}
        \caption{Self-reflection prompt}
        \label{fig:reflection}
    \end{minipage}
    \hfill
    \begin{minipage}[b]{0.327\textwidth}
        \includegraphics[width=\textwidth]{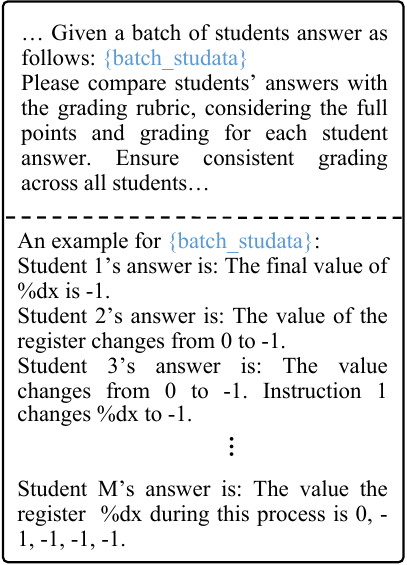}
        \caption{Batching prompt}
        \label{fig:batching}
    \end{minipage}
\end{figure}

\paragraph{Prompt Strategy}
The effectiveness of grading is directly influenced by the quality of the prompts. Therefore, at this stage, we implement three different prompt strategies: \textit{one-shot prompts}, \textit{self-reflection prompts}, and \textit{batching prompts}.
In Section~\ref{s:eval_prompts}, we will explore the impact of these prompt strategies on grading performance through experiments.
We now introduce each of these prompt strategies respectively as follows. 
\begin{itemize}
    \item \textit{One-Shot Prompts:} LLMs have a good ability for in-context learning. Therefore, when generating prompts, we provide an additional example for the LLM to learn from. Figure~\ref{fig:oneshot} shows an example of such a prompt.
    \item \textit{Self-Reflection Prompts:} In this prompting strategy, each student's answer is initially graded by the LLM. This is followed by several iterations where the LLM is asked to reflect on its previous output and provide a more robust score and rationale. Figure~\ref{fig:reflection} illustrates this strategy's workflow and provides a sample instruction for self-reflection.
    \item \textit{Batching Prompts:} In this prompting strategy, several students' answers are processed in batches and graded together by the LLM. By considering multiple students' answers jointly, the LLM can better maintain fairness and consistency across them. Figure~\ref{fig:batching} shows an example of such a prompt.
\end{itemize}

\subsection{Post-grading Review}
\label{s:score_review}

LLMs may produce unreasonable grading results due to randomness or hallucination issues. Therefore, a post-grading review of all results is conducted after all student responses have been scored. In this section, we detail the review method.

We adopt a group comparison method to review the grading results and identify all unreasonable items. 
Considering the question being graded $q$, the grade rubric $r$, and the set of $n$ students' answers $A = \{a_1, \ldots, a_n\}$. After the previous stages, we obtain a set of answer-score pairs, forming the set $D = \{(a_i, g_i)\}$.
Given a group size $c$, we randomly divide the elements of $D$ into $n/c$ groups, with each group $D_i = \{(a_{i,1}, g_{i,1}), \ldots, (a_{i,c}, g_{i,c})\}$. Our goal is to use an LLM to identify all anomalies within each group $D_i$.
Specifically, the LLM will be asked to check:
1) For each $(a_i, g_i)$, whether $g_i$ deviates significantly from the requirements of the scoring rubric $r$;
2) Whether there are significant inconsistencies among the scores of these $c$ answers.
The LLM will return all detected anomalies. This process can be represented as:
\begin{equation}
    \text{outlier} = E(q, r, p, D_i)
\end{equation}
where $p$ contains the system prompt and other necessary context and instructions.
The value of the group size $c$ depends on ensuring that batching these $c$ pairs together does not exceed the context length of the LLM.

\begin{wrapfigure}{r}{0.5\textwidth}
%\begin{figure}
    \centering
    \includegraphics[width=0.48\textwidth]{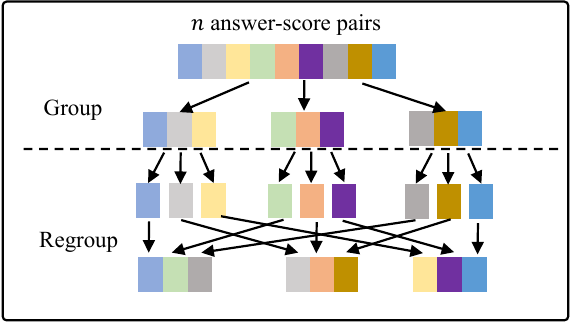}
    \caption{Our proposed re-grouping strategy.}
    \label{fig:regroup}
%\end{figure}
\end{wrapfigure}

\paragraph{Re-Grouping Strategy}
Intuitively, by comparing each grading result with a wider variety of other results, we can further improve the accuracy of the review. Therefore, we propose a re-grouping strategy, which is illustrated in Figure~\ref{fig:regroup}. 
During the re-grouping process, each group is further divided into $k$ sub-groups. From each sub-group, a set of sub-groups is sampled and recombined to form new groups. These newly formed sub-groups are then sent to the LLM for review. 
This re-grouping strategy enables a greater mixture of grading results, which helps in identifying outliers while maintaining the overall data distribution characteristics of $n$, despite the limitations of token length.

\section{\ourdataset~Dataset}
\label{s:os_dataset}

\begin{figure}[] 
\centering
\small
\begin{tabularx}{\textwidth}{Xc}
    \hline
    \multicolumn{2}{p{13.5cm}}{\centering\textbf{Question Description}} \\ % \hline
    \multicolumn{2}{p{13.5cm}}{\textbf{(Full points: 15)} Now let’s look at \verb|vector-avoid-hold-and-wait.c|. What is the main problem with this approach? How does its performance compare to the other versions (i.e., \verb|vector-global-order| and \verb|vector-try-wait|), when running both with \verb|-p| and without it?} \\ \hline
    \multicolumn{2}{p{13.5cm}}{\centering\textbf{Standard Answer \& Grading Criteria}} \\ % \hline
    \multicolumn{2}{p{13.5cm}}{
        (1) The main problem with this approach is that it is too coarse ...
        % : the global lock (which protects the acquisition of all the other locks) will be under contention even when the vectors being manipulated by each thread is different.  
        \textbf{(\textit{5 points})}
        (2) The run times under heavy contention ...
        % (without the \verb|-p| flag) very close to those for vector-global-order. 
        \textbf{(\textit{5 points})}
        (3) For the high parallelism (with the \verb|-p| flag) case, ...
        % \verb|vector-avoid-hold-and-wait| performs as it does under high contention, but about twice as slow as \verb|vector-global-order| under high parallelism. 
        \textbf{(\textit{5 points})}
    } \\ \hline % \hline
    \multicolumn{1}{p{10cm}}{\centering\textbf{Student Answers}} & {\centering\textbf{Human-Given Grades}} \\  % \hline
    \multicolumn{1}{p{10cm}}{The main problem is that the global lock holds the lock even when sometimes it is not necessary to avoid deadlock, it plays too safe. The performance ...
    % without \verb|-p|: global order better than avoid hold and wait better than try wait. But with \verb|-p|: global order better than try wait better than avoid hold and wait.
} & \makecell[c]{13, 10, 10} \\ % \hline
    \multicolumn{1}{p{10cm}}{The main problem with this approach is that it is too coarse: the global lock (which protects the acquisition of all the other locks) will ...
    % be under contention even when the vectors being manipulated by each thread is different. The run times under heavy contention (without the \verb|-p| flag) very close to those for \verb|vector-global-order|. For the high parallelism (with the \verb|-p| flag) case, \verb|vector-avoid-hold-and-wait| performs as it does under high contention, but about twice as slow as \verb|vector-global-order| under high parallelism. Consequently, the comparison between vector-global-order and \verb|vector-try-wait| holds here as well.
    } & 15, 13, 15 \\ \hline
\end{tabularx}
\caption{An example from our OS dataset.
% with a question, the correct answer, and two student answers.
%The code involved in the question can be accessed and run by students on the server of the simulate environment. 
}
\label{fig:example_os}
\end{figure}

We have collected and open-sourced a dataset named \ourdataset, aimed at assisting researchers in evaluating the capabilities of LLMs in grading tasks. 
The dataset is sourced from the Operating Systems course of undergraduate students at Computer Science department of City University of Hong Kong.
%\textit{Anonymous University}. 
The dataset includes 6 questions, selected from the course's tutorials and assignments. 
Figure \ref{fig:example_os} shows one question from the dataset, along with answers from two students.
Each question contains a detailed description and necessary supplementary materials (such as executable code), along with the total score value, standard answers, and detailed grading criteria.
For each question, we collected answers from 40 students. These student answers were human-graded, with each answer being evaluated by three teaching assistants. 
The scores ranged from 0 to the maximum score for that question. 
Ultimately, the average of the three human scores was used as the student’s final score. 
% Figure \ref{fig:example_os} shows a question-answer pair from the dataset, along with example answers from two students.

% We create a data set of questions from Operating System assignments with answers provided by a class of undergraduate students. The assignments were administered as part of a Operating System course at the \textit{Anonymous University}. 
% For each assignment, the student answer sheets were collected via online system. The student submitted answers to 6 questions spread across 7 tutorials and 2 assignments. Figure \ref{fig:example_os} shows a question-answer pairs with two sample student answers each. 50 students were enrolled in the class and submitted answers to these assignments. The data set we work with consisted of a total of 240 student answers. 
% % This is less than the expected $50 \times 6 = 300$  as some students did not submit answers for a few assignments. 

% The answers were independently graded by three human graders, using a score from 0 (completely incorrect) to the \emph{full\_points} (perfect answer) for that question (\emph{full\_points} could vary across different questions), with a minimum granularity of 0.5 points. All human graders were PhD students in the computer science department and were teaching assistants for the Operating Systems course, and grader 3 was one of the authors of this paper. We treat the average grade of the three annotators as the gold standard against which we compare the output of our system.

\section{Experiments}
\label{s:exp}

In this section, we implement our system and test it using the \ourdataset~dataset mentioned in Section \ref{s:os_dataset}, as well as an open-source dataset, Mohler~\citep{mohler2011learning,mohler2009text}. 
The Mohler dataset is a question-answer grading database in the ASAG field, with questions pertaining to computer science. The organization of the Mohlar dataset is identical to the \ourdataset~dataset shown in Figure \ref{fig:example_os}, but it does not include grading criteria. A more detailed introduction can be found in Appendix~\ref{appendix:mohler}.
Our system is deployed on a server equipped with two A100 GPUs and 400GB of memory.
When implementing our system, we utilized OpenAI's API~\cite{openai-api} as our LLM source, and employed LangChain~\cite{langchain} to implement the agents and workflows. 
To objectively evaluate our system's performance in the grading task, we used multiple evaluation metrics, including MAE and RMSE, among others. These metrics are detailed in Appendix~\ref{Appendix:eval_metric}. 
Since the outputs of LLMs inherently possess uncertainty, during experiments, each metric is measured multiple times and the average value is presented as the result.

\subsection{Evaluations of Various Generated Rubrics}

%%%table 1
\begin{table}[]
\centering
\caption{The grading performance under different rubric generation methods for the \ourdataset~dataset}
\label{tab-gencri}
\resizebox{\textwidth}{!}{%
\begin{tabular}{rccccccccc}
\hline
\rule{0pt}{1.1em} % 设置行高
 \textbf{Methods} & \textbf{MAE} $\downarrow$ & \textbf{NRMSE} $\downarrow$ & \textbf{RMSE} $\downarrow$ & \textbf{Perason} $\uparrow$ &  & \textbf{MAE} $\downarrow$ & \textbf{NRMSE} $\downarrow$ & \textbf{RMSE} $\downarrow$ & \textbf{Perason} $\uparrow$ \\ \hline   \hline
 & \multicolumn{4}{c}{\textit{Q1}} &  & \multicolumn{4}{c}{\textit{Q2}} \\ \cline{2-5} \cline{7-10} 
\textbf{Coarse-gr., TA} & 6.97 & 8.98 & 0.47 & 0.46 &  & 5.90 & 7.57 & \textbf{0.47} & 0.10 \\
\textbf{Fine-gr., TA} &  4.51 & 5.62 & 0.30 & 0.74 &  & 5.90 & 7.75 & 0.48 & 0.15 \\
\textbf{Gen., Rand.} & 4.42 & 5.65 & 0.30 & 0.75 &  & 5.95 & 7.65 & 0.48 & 0.32 \\
\textbf{Gen., Distr.} & \textbf{4.05} & \textbf{5.21} & \textbf{0.27} & \textbf{0.76} &  & \textbf{5.70}  & \textbf{7.46}  & \textbf{0.47} & \textbf{0.37} \\ % \hline % \hline
 & \multicolumn{4}{c}{\textit{Q3}} &  & \multicolumn{4}{c}{\textit{Q4}} \\  \cline{2-5} \cline{7-10} 
\textbf{Coarse-gr., TA} & 2.85 & 3.41 & 0.23 & 0.62 &  & 5.45 & 7.32 & 0.46 & 0.36 \\
\textbf{Fine-gr., TA} & 2.40 & 3.38 & 0.23 & 0.71 &  & 4.78 & 6.25 & 0.39 & 0.45 \\
\textbf{Gen., Rand.} & 2.50 & 3.50 & 0.23 & 0.72 &  & 4.10 & 5.83 & 0.36 & 0.55 \\
\textbf{Gen., Distr.} & \textbf{2.35} & \textbf{3.17} & \textbf{0.21} & \textbf{0.75} &  & \textbf{3.38} & \textbf{4.89}  & \textbf{0.31} & \textbf{0.60} \\ % \hline  % \hline
 & \multicolumn{4}{c}{\textit{Q5}} &  & \multicolumn{4}{c}{\textit{Q6}} \\  \cline{2-5} \cline{7-10} 
\textbf{Coarse-gr., TA} & 10.90 & 13.10 & 0.49 & 0.53 &  & 9.22 & 12.86 & 0.32 & 0.42 \\
\textbf{Fine-gr., TA} & 9.62 & 11.77 & 0.44 & 0.49 &  & 10.55 & 12.83 & 0.32 & 0.47 \\
\textbf{Gen., Rand.} & 6.05 & 8.41 & 0.31 & 0.64 &  & 9.62 & 12.62 & 0.32 & 0.40 \\
\textbf{Gen., Distr.} & \textbf{6.03} & \textbf{8.21} & \textbf{0.30}  & \textbf{0.66} &  & \textbf{8.72} & \textbf{11.42} & \textbf{0.29} & \textbf{0.50} \\ \hline
\end{tabular}%
}
\end{table}
%% table 2
\begin{table}[]
\centering
\caption{The grading performance under different rubric generation methods for the Mohler dataset}
\label{tab-gencri-mohler}
\resizebox{0.96\textwidth}{!}{%
\begin{tabular}{rccccccccc}
\hline
\rule{0pt}{1.1em} % 设置行高
 \textbf{Methods} & \textbf{MAE} $\downarrow$ & \textbf{NRMSE} $\downarrow$ & \textbf{RMSE} $\downarrow$ & \textbf{Perason} $\uparrow$ &  & \textbf{MAE} $\downarrow$ & \textbf{NRMSE} $\downarrow$ & \textbf{RMSE} $\downarrow$ & \textbf{Perason} $\uparrow$ \\ \hline  \hline
 & \multicolumn{4}{c}{\textit{Q1.1}} &  & \multicolumn{4}{c}{\textit{Q2.6}} \\  \cline{2-5} \cline{7-10} 
\textbf{Coarse-gr.} & \textbf{0.78} & \textbf{1.02} & \textbf{0.29} & \textbf{0.69} & & \textbf{0.90} & \textbf{1.11} & \textbf{0.37} & \textbf{0.34}   \\
\textbf{Fine-gr.} & 1.67  & 2.06 & 0.59 & 0.03 & & 1.83 &  2.00 & 0.67 & 0.37 \\ % \hline % \hline
 & \multicolumn{4}{c}{\textit{Q2.7}} &  & \multicolumn{4}{c}{\textit{Q3.6}} \\  \cline{2-5} \cline{7-10} 
\textbf{Coarse-gr.} & \textbf{0.59} & \textbf{0.74} & \textbf{0.15} & \textbf{0.65} &  & \textbf{0.43} & \textbf{0.87} & \textbf{0.35} & \textbf{0.60} \\
\textbf{Fine-gr.} & 0.90 & 1.11 & 0.22 & 0.55 &  & 0.61 & 1.21 & 0.45 & 0.53 \\%  \hline % \hline
 & \multicolumn{4}{c}{\textit{Q4.2}} &  & \multicolumn{4}{c}{\textit{Q11.7}} \\  \cline{2-5} \cline{7-10} 
\textbf{Coarse-gr.} & \textbf{0.57} & \textbf{0.80} & \textbf{0.27} & \textbf{0.68} &  & \textbf{0.68} & \textbf{0.80} & \textbf{0.28} & \textbf{0.74} \\
\textbf{Fine-gr.} & 0.50 & 0.89 & 0.30 & 0.64 &  & 0.72 & 0.97 & 0.31 & 0.68 \\ \hline
\end{tabular}%
}
\end{table}

We first evaluate the impact of different rubrics on the grading effectiveness of LLMs and simultaneously verify the effectiveness of the algorithm proposed in Section~\ref{s-gencri}.
Table~\ref{tab-gencri} presents the evaluation results on the \ourdataset~dataset, while Table~\ref{tab-gencri-mohler} shows the evaluation results on the Mohler dataset.

The evaluation on the \ourdataset~dataset includes four different grading rubrics. Among them, the \textbf{Coarse-gr., TA} and \textbf{Fine-gr., TA} rubrics are both manually generated rubrics that we followed when scoring student answers. 
The difference is the granularity: The former provides only the standard answer, while the latter additionally includes a grading criteria, as illustrated in Figure~\ref{fig:example_os}. \textbf{Gen., Rand.} and \textbf{Gen., Distr.} are respectively derived from the two rubric generation algorithms we demonstrated in Section~\ref{s-gencri}.
The rubrics for the Mohler dataset differ. Since the Mohler dataset does not provide grading criteria, it includes only two rubrics: \textbf{Coarse-gr.}, which directly comes from the standard answer for each question included in the dataset, and \textbf{Fine-gr.}, which is generated using the rubric generation algorithm.
In Appendix~\ref{appendix:rubric}, we present several examples to illustrate the different rubrics.

% \paragraph{Rubric Granularity v.s. Performance}
We compared the performance using different granularities of rubrics. In Tables~\ref{tab-gencri} and~\ref{tab-gencri-mohler}, we observed completely opposite phenomena: for the \ourdataset~dataset, fine-grained rubrics significantly improved scoring performance, whereas for the Mohler dataset, fine-grained rubrics actually reduced scoring performance.
We attribute it to the differing difficulties of the two datasets. 
For the OS dataset, each question is relatively long and complex, involving multiple steps and often including code execution. For such problems, LLMs lack sufficient capability to provide correct answers on their own, so detailed rubrics help LLMs score correctly.
In contrast, the questions in the Mohler dataset are relatively simple. For example, one question in the Mohler dataset is: \textit{``What is the role of a prototype program in problem solving?''}
For such questions, the world knowledge of LLMs is sufficient to provide good answers, as well as to evaluate students' responses. Therefore, not overly constraining the LLMs can better leverage their abilities~\cite{patil2023gorilla}.
Additionally, in Table~\ref{tab-gencri}, we further observe that our rubric generation algorithm can further improve the grading performance on complex questions.
Based on this, we claim that our approach is effective for complex problems and further demonstrate the importance of our \ourdataset~dataset for better evaluating the performance of LLMs on grading tasks.
Therefore, in the subsequent evaluation, we mainly focus on the improvements our method brings to the \ourdataset~dataset.

\begin{table}[t]
\centering
\caption{The impact of different prompt strategies on grading performance for the \ourdataset~dataset}
\label{tab-prompts}
\resizebox{\textwidth}{!}{%
\begin{tabular}{rccccccccc}
\hline
\rule{0pt}{1.1em} % 设置行高
 \textbf{Methods} & \textbf{MAE} $\downarrow$ & \textbf{NRMSE} $\downarrow$ & \textbf{RMSE} $\downarrow$ & \textbf{Perason} $\uparrow$ &  & \textbf{MAE} $\downarrow$ & \textbf{NRMSE} $\downarrow$ & \textbf{RMSE} $\downarrow$ & \textbf{Perason} $\uparrow$ \\ \hline   \hline
 & \multicolumn{4}{c}{\textit{Q1, batch size=10}} &  & \multicolumn{4}{c}{\textit{Q2, batch size=40}} \\ \cline{2-5} \cline{7-10} 
\textbf{Baseline} & 4.28 & 5.81 & 0.31 & 0.73 &  & 7.35 & 9.58 & 0.60 & 0.30 \\
\textbf{One-shot} &  3.74 & 5.33& 0.28 & 0.77 &  & 5.70 & 7.46 & 0.47 & 0.37 \\
\textbf{Self-reflection} & 3.83 & 4.98 & 0.26 & 0.76 &  & 5.95 & 7.68 & 0.48 & 0.36 \\
\textbf{Batching} & \textbf{3.59} & \textbf{4.64} & \textbf{0.24} & \textbf{0.79} & & \textbf{3.09} & \textbf{6.28} & \textbf{0.39} & \textbf{0.42} \\ % \hline % \hline
 & \multicolumn{4}{c}{\textit{Q3, batch size=40}} &  & \multicolumn{4}{c}{\textit{Q4, batch size=30}} \\  \cline{2-5} \cline{7-10} 
\textbf{Baseline} & 2.00 & 2.89 & 0.19 & 0.74 &  & 3.33 & 5.48 & 0.34 & 0.46 \\
\textbf{One-shot} &  \textbf{1.95} & \textbf{2.72} & \textbf{0.18} & \textbf{0.79} &  & \textbf{3.23} & \textbf{4.81} & \textbf{0.30} & \textbf{0.59} \\
\textbf{Self-reflection} &  2.55 &  3.21 & 0.21 & 0.74 &  & 3.85 & 5.55 &  0.35 &0.42 \\
\textbf{Batching} & 2.98 & 3.95 & 0.26 & 0.44 & & 3.83 & 5.93 & 0.37 & 0.32 \\ % \hline % \hline
 & \multicolumn{4}{c}{\textit{Q5, batch size=40}} &  & \multicolumn{4}{c}{\textit{Q6, batch size=20}} \\  \cline{2-5} \cline{7-10} 
 \textbf{Baseline} & 6.00 & 8.07 & 0.30 & 0.66 &  & 11.00 & 14.6 & 0.36 & \textbf{0.22} \\
\textbf{One-shot} &  \textbf{4.15} & \textbf{6.07} & \textbf{0.22} & \textbf{0.78} &  & \textbf{7.95} & \textbf{10.23} & \textbf{0.26} & \textbf{0.22} \\
\textbf{Self-reflection} & 7.33 & 9.46 & 0.35 & 0.57 &  & 10.90 & 14.53 & 0.36 & \textbf{0.22} \\
\textbf{Batching} & 8.20 & 10.33 & 0.38 & 0.63 & & 11.91 & 16.45 & 0.41 & 0.21 \\ \hline 
\end{tabular}%
}
\end{table}

\subsection{Evaluations of Prompt Strategies in Grading}
\label{s:eval_prompts}

In this section, we assess the impact of different prompting strategies on performance during the grading process. The evaluation results are presented in Table~\ref{tab-prompts}. The \emph{batch size} in the table corresponds to the setting of the Batching prompts.
For all questions in the \ourdataset~dataset, our prompting strategies exceeded the baseline levels. 
The results indicate that different questions have different optimal prompting strategies, with Batching and One-Shot performing the best among all strategies. We also observed that, in certain cases, some strategies could reduce the grading performance. These observations suggest that it is crucial to design different prompting strategies tailored to the characteristics of specific questions.

\subsection{Evaluations for Score Review}

In this section, we evaluate the impact of the review process on grading performance and the effectiveness of re-grouping strategies in improving the accuracy of identifying unreasonable results.

\begin{wraptable}{r}{0.55\textwidth}
    \centering
    \caption{Evaluation of re-grouping strategy}
    \label{tab-regroup}
    \resizebox{.55\textwidth}{!}{%
        \begin{tabular}{rcccccc}
        \hline
         & \multicolumn{6}{c}{\textbf{Accuracy $\uparrow$}} \\ \cline{2-7} 
        \textbf{Methods} & \textit{Q1} & \textit{Q2} & \textit{Q3} & \textit{Q4} & \textit{Q5} & \textit{Q6} \\ \hline
        \textbf{Baseline} & 0.58 & 0.64 & 0.59 & 0.72 & 0.55 & 0.70 \\
        \textbf{Re-grouping} & \textbf{0.62} & \textbf{0.73} & \textbf{0.71} & \textbf{0.77} & \textbf{0.64} & \textbf{0.76} \\ \hline
        \end{tabular}
    }
\end{wraptable}

% In this section, we evaluate the impact of the review process on grading performance and the effectiveness of re-grouping strategies in improving the accuracy of identifying unreasonable results. 
%
During the testing process, we randomly selected some students and modified their scores to simulate the occurrence of unreasonable results. The evaluation results are presented in Table~\ref{tab-regroup}.
The results show that through the review process, the LLM can detect anomalies in the grading, thereby further improving the accuracy. Additionally, the re-grouping strategy can further enhance the model's accuracy.

\section{Limitations}
\label{s:limit}
\iffalse
本文提出一种新的基于multi-agent的grading system. But we want to highlight that there are remain som limitations or improvement spaces: 
1) 本文grade的问题类型是computer science领域的。实际grading task所涉及的题目类型比较广泛，比如essay scoring, summarization evaluation等，对于不同类型的具体grading task， 我们提出的系统方法的潜能还需要探究，这也是我们未来research关注的方向。
2) \textbf{Time Efficiency} poses a problem in our framework. To build a such a collaborative system (i.e., multi-agent) with grading task, it requires multiple interations with LLMs throughout the whole workflow and thus brings increasing time costs for generating the response;
3) \textbf{Token Const} is another common problem when using LLM, since the maxinum token length is always limited. Although some works have extended the maxinum length to 128K, it is still insatiable for us if we want to grading大量的学生作业。 Therefore, how to brively and effectively summarize model descriptions is also worthy of exploration;

token cost是使用LLM做grading的另一个重大问题。随着批改的作业量增加，token cost也会随之增加。Therefore, finding ways to concisely and effectively summarize model descriptions is a valuable area for exploration. 

\fi

This paper proposes a novel multi-agent grading system. However, we want to highlight several limitations and areas for improvement: 1) \textbf{Domain Specificity}: The datasets used in this study primarily focus on questions in a special domain. Exploring its applicability to grading tasks in other domains remains an important avenue for future research. 2) \textbf{Time Efficiency}: Time efficiency poses a significant challenge in our framework. Building a collaborative system (i.e., multi-agent) for grading tasks requires multiple interactions with LLMs throughout the entire workflow, resulting in increased time costs for generating responses.
3) \textbf{Token Cost}: Token cost is another common issue when utilizing LLMs for grading. As the volume of answers to be graded escalates, the token cost correspondingly increases. Therefore, developing strategies to concisely and effectively summarize model descriptions presents a crucial area for exploration.
% Token cost is another common issue when using LLMs, as the maximum token length is always limited. Although some works have extended the maximum length to 128K, it is still insufficient for grading a large volume of student assignments. Therefore, finding ways to concisely and effectively summarize model descriptions is a valuable area for exploration. 

\section{Conclusion}
\label{s:conclusion}

This paper introduces ``\oursystem,'' a multi-agent grading system.
Current automated short answer grading (ASAG) systems still face challenges in terms of accuracy, consistency, and fairness. Drawing inspiration from human grading practices, we propose a systematic approach that divides the grading task into three stages: rubric generation, grading, and review. Each stage is designed with specialized agents to enhance performance.
Additionally, we collected and open-sourced an \ourdataset~dataset from an undergraduate operating systems course.
We implemented the system and evaluated it on both the \ourdataset~dataset and the Mohler dataset. Experimental results show significant performance improvements with our system.

\bibliography{reference}

%%%%%%%%%%%%%%%%%%%%%%%%%%%%%%%%%%%%%%%%%%%%%%%%%%%%%%%%%%%%
\appendix
\newpage
\section{Appendix}
\label{s:appendix}

% %%%%%%% 
\subsection{Prompts for Rubric Generation}
\label{appedix:prompt_generate_rurbric} 

\begin{figure}[h]
    \centering
    \includegraphics[width=0.6\textwidth]{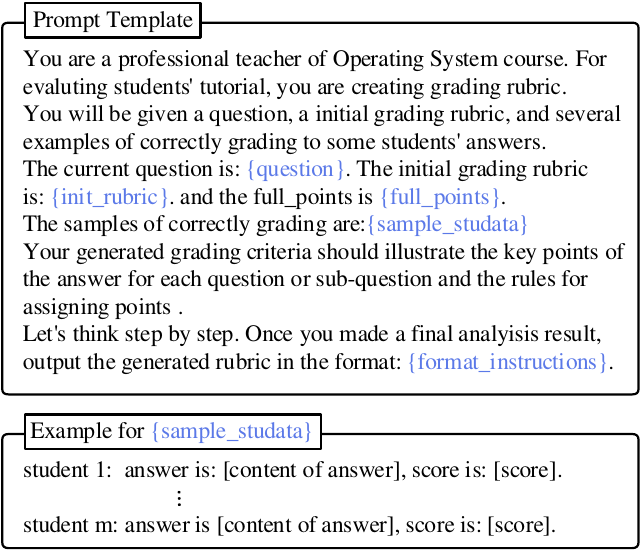}
    \caption{Example of prompts for rubric generation.}
    \label{fig:prompt_genri}
\end{figure}

Figure \ref{fig:prompt_genri} shows the prompt remplate for LLMs to generated rubric. The prompt consists of the instruction, all input data (Blue color part in Figure \ref{fig:prompt_genri}), and the variable ${sample\_studata}$ is the sampled student answer-score pairs. The generated rubric contains reference answer and detailed scoring rules.

%%%%%%
\subsection{Mohler Dataset}
\label{appendix:mohler}
 % Mohler dataset comprises question-and-answer pairs from a Computer Science course. The dataset was constructed by collecting answers from 31 students across 10 assignments and 2 exams conducted as part of the course. It contains a total of 2,273 student answers associated with 80 questions. The assignment answers were evaluated by two human graders who assigned scores ranging from 0 to 5. The exam question answers were scored on a scale of 0 to 10, and the scores were normalized to align with the score range of the assignment answers. The dataset is publicly available, and each entry includes the question ID, the question itself, the student's answer, the desired answer, and the two evaluation scores. Figure~\ref{fig:example_mohler} show the example from Mohler dataset.

The Mohler dataset consists of question-and-answer pairs from a Computer Science course. It was constructed by collecting responses from 31 students across 10 assignments and 2 exams. In total, the dataset includes 2,273 student answers corresponding to 80 questions. 
For the assignments, two human graders evaluated the answers, assigning scores ranging from 0 to 5. Exam question answers were scored on a scale of 0 to 10, and these scores were subsequently normalized to match the assignment scoring range. Each entry in the dataset includes the question ID, the question itself, the student's answer, the desired answer, and the two evaluation scores.
The dataset is publicly available, providing a valuable resource for research in automated grading. Figure~\ref{fig:example_mohler} shows an example from the Mohler dataset.

 \begin{figure}[h] 
\centering
\small
\begin{tabularx}{\textwidth}{Xc}
    \hline
    \multicolumn{2}{p{13.5cm}}{\centering\textbf{Question Description}} \\ % \hline
    \multicolumn{2}{p{13.5cm}}{\textbf{(Full points: 5)} What is the role of a prototype program in problem solving?} \\ \hline
    \multicolumn{2}{p{13.5cm}}{\centering\textbf{Standard Answer \& Grading Criteria}} \\ % \hline
    \multicolumn{2}{p{13.5cm}}{
        To simulate the behaviour of portions of the desired software product. \textbf{(\textit{5 points})}
    } \\ \hline % \hline
    \multicolumn{1}{p{10cm}}{\centering\textbf{Student Answers}} & {\centering\textbf{Human-Given Grades}} \\  % \hline
    \multicolumn{1}{p{10cm}}{To simulate portions of the desired final product with a quick and easy program that does a small specific job. It is a way to help see what the problem is and how you may solve it in the final project.
} & \makecell[c]{5, 5} \\ % \hline
    \multicolumn{1}{p{10cm}}{A prototype program simulates the behaviors of portions of the desired software product to allow for error checking.
    } & 5, 3 \\ \hline
\end{tabularx}
\caption{An example from Mohler dataset.}
\label{fig:example_mohler}
\end{figure}

% %%%%%%% 
\subsection{Detailed Implementations of Sampling Algorithms}
\label{appedix:sampling}
Algorithm~\ref{alg_random} and Algorithm~\ref{alg_distribute} are detailed implementations of algorithms for generating rubrics based on the sampling method.

% algorithm 1
\begin{algorithm}[]
\small
\renewcommand{\algorithmicrequire}{\textbf{Input:}}
\renewcommand{\algorithmicensure}{\textbf{Output:}}
    \caption{Rubric generation based on \emph{Random} sampling method}
    \label{alg_random}
    \begin{algorithmic}[1]
        \Require For question \(q\), Current rubric \(r_i\), Student answers set \( A = \{a_1, a_2, ..., a_n\} \), Sample size $m$
        \Ensure Improved rubric \(r_{i+1}\)
        \State Initialize $\Tilde{A_m} = \emptyset$
        \State $\Tilde{A_{m}} \gets$ RandomSample$(A, m)$
        \State $A_{m}=$ HumanLabel$(\Tilde{A_m})$, where $A_{m}=\{(a_i, g_i), i\in (1,m)\}$, $g_i$ is the human score
        \State $p \gets$ PromptConstruct$(q, r_i, A_{m})$
        \State $r_{i+1}=$ LLMGenerate($p$)
        \State \textbf{return} $r_{i+1}$
    \end{algorithmic}
\end{algorithm}
%%
% algorithm 2
\begin{algorithm}[]
\small
\renewcommand{\algorithmicrequire}{\textbf{Input:}}
\renewcommand{\algorithmicensure}{\textbf{Output:}}
\caption{Rubric generation based on \emph{Distribution-Aware} sampling method}
\label{alg_distribute}
\begin{algorithmic}[1]
\Require Student answers set $A = \{a_1, a_2, \ldots, a_n\}$, Current rubric $r_i$, Sample size $m$
\Ensure Improved rubric \(r_{i+1}\)
\State $D=$ LLMGrate$(q,r_i,A)$, Where $D=\{(a_i,\hat{g_i}), i \in (1,n)$\}, $\hat{g_i}$ is the grading score of LLM based on the current rubric $r_i$
\State Calculate the distribution of scores and divide $D$ into $k$ strata $\{B_1, B_2, \ldots, B_k\}$
\For{$l = 1$ to $k$}
    \State Compute the proportion $p_l = |B_l| / n$
    \State Determine the number of samples for stratum $B_l$ as $m_l = \lceil p_l \times m \rceil$
\EndFor
\State Initialize $\Tilde{A_m} = \emptyset$
\For{$l = 1$ to $k$}
    \State Randomly sample $m_l$ answers from $B_l$
    \State $\Tilde{A_m} = \Tilde{A_m} \cup \{\text{selected samples from } B_l\}$
\EndFor
\State $A_{m}=$ HumanLabel$(\Tilde{A_m})$, where $A_{m}=\{(s_i, g_i), i\in (1,m)\}$, $g_i$ is the human score
\State $p \gets$ PromptConstruct$(q, r_i, A_{m})$
\State $r_{i+1}=$ LLMGenerate($p$)
\State \textbf{return} $r_{i+1}$
\end{algorithmic}
\end{algorithm}

%%%%%%%
\subsection{Examples of Various Rubric Types}
\label{appendix:rubric}
\iffalse
Typically, the grading rubric for one question consists of multiple sub-questions有sample answer和grading criteria组成， sample answer是问题的答案内容， grading criteria分配分数的原则指导，形式如“If the answer mention X, then it receives a score of Y”. 
Figure 8 展示了针对同一个问题的四种不同类型的Rubric。 
1) Coarse-grained rubric by TA 是由Teacher制定的粗粒度的rubric, 只提供了本道题的标准答案内容。 2） Fine-grained rubric by TA 在sample answer的基础上又提供了grading critiera, 是指导分数分配的规则。3）Generated Rubric by random algorithm的包含sample answer和grading criteira, 相比较ine-grained rubric by TA 只是分配了每个sub-question的分值，还增加了一些answer case应该被分配的分数范围。4）Generated rubric by distribute algorithm的format形式与generated rubric by random相似，区别在于给出的一些answer case内容以及被分配的分数范围不同。
\fi

Typically, the grading rubric for a question consists of multiple sub-questions, each with a sample answer and associated grading criteria. The sample answer provides the expected content of the response, while the grading criteria outline the principles for assigning scores, often in the form of `` If the answer mentions X, then it receives a score of Y. '' Figure~\ref{fig:example_rubric}  illustrates four different types of rubrics for the same question: 1) Coarse-grained Rubric by TA: This rubric, created by Teachers (TA), is coarse-grained and only provides the standard answer content for the question.
2) Fine-grained Rubric by TA: In addition to the sample answer, this rubric includes grading criteria, offering rules for score allocation.
3) Generated Rubric by Random Algorithm: This rubric includes both sample answers and grading criteria. Compared to the fine-grained rubric by the TA, it not only assigns scores to each sub-question but also adds score ranges for specific answer cases.
4) Generated Rubric by Distribute Algorithm: The format of this rubric is similar to that of the rubric generated by the random algorithm. However, the content of some answer cases and the assigned score ranges differ.

\begin{figure}[] 
\centering
\small
\begin{tabularx}{\textwidth}{Xc}
    \hline
    \multicolumn{2}{p{13.5cm}}{\centering\textbf{Question Description}} \\ % \hline
    \multicolumn{2}{p{13.5cm}}{\textbf{(Full points: 19)} Now do the same but with jobs of different lengths: 100, 200, and 300. The commands are (\verb|./scheduler.py -p SJF -l 100,200,300|) and (\verb|./scheduler.py -p FIFO -l 100,200,300|). Use the -c flag to check your answers. What if you change the order of the job length? Try different orders to find the difference. 
    } \\ \hline
    \multicolumn{2}{p{13.5cm}}{\centering\textbf{Sample Answer}} \\ % \hline
    \multicolumn{2}{p{13.5cm}}{
        For SJF and FIFO:
        
        Job   0 -- Response: 0.00    Turnaround 100.00  Wait 0.00 
        
        Job   1 -- Response: 100.00  Turnaround 300.00  Wait 100.00 
        
        Job   2 -- Response: 300.00  Turnaround 600.00  Wait 300.00
        
        Average -- Response: 133.33  Turnaround 333.33  Wait 133.33 
        
        The response time and turnaround time for three jobs using the FIFO policy and SJF policy are the same in the given order of job length. If we change the order of job length, response time and turnaround time for the SJF policy should remain the same. 
        However, the results for FIFO policy in different orders will change. In particular, if we run these jobs in order 300, 200, 100, it makes FIFO perform poorly (the average response time is 266.67, and turnaround time is 466.67).  
    } \\ \hline % \hline
    \multicolumn{2}{p{13.5cm}}{\centering\textbf{Coarse-grained Rubric, TA}}  \\  % \hline
    \multicolumn{2}{p{13.5cm}}{The sample answer is : \textbf{[Sample Answer]} 
    }  \\  \hline
    \multicolumn{2}{p{13.5cm}}{\centering\textbf{Fine-grained Rubric, TA}}  \\  % \hline
    \multicolumn{1}{p{13.5cm}}{
    The sample answer is : \textbf{[Sample Answer]}
    
    The grading criteria is :
    
      1. For correctly answer “Average response time and Turnaround time” for SJF (6.5 points)
      
      2. For correctly answer “Average response time and Turnaround time” for FIFO (6.5 points)
      
      3. For pointing out “if change the order of job length, response time and turnaround time for the SJF policy should remain the same.” (3 points)
      
      4. For correctly explaining “the results for FIFO policy in different orders will change.” (3 points)
    } \\ \hline
    \multicolumn{2}{p{13.5cm}}{\centering\textbf{Generated Rubric, Random}}  \\  % \hline
    \multicolumn{2}{p{13.5cm}}{
    The sample answer is : \textbf{[Sample Answer]}
    
    The grading criteria is :
    
     1. For correctly calculating the average response time and turnaround time for SJF (6.5 points) 
     
     2. For correctly calculating the average response time and turnaround time for FIFO (6.5 points) 
     
     3. For identifying that changing the order of job length does not affect the response time and turnaround time for the SJF policy (3 points) 
     
    4. For explaining that the results for FIFO policy will change when the order of job length is changed (3 points) 
    
    5. For providing detailed and accurate explanations of how SJF and FIFO policies work in different job length orders (up to 6 points) 
    
    6. For demonstrating a clear understanding of how SJF and FIFO policies handle job scheduling in different scenarios (up to 6 points) 
    
    7. For accurately comparing the response time and turnaround time results between SJF and FIFO policies in different job length orders (up to 6 points) 
    
    8. For providing a well-structured and coherent explanation of the differences in response time and turnaround time between SJF and FIFO policies (up to 6 points)
    } \\ \hline
    \multicolumn{2}{p{13.5cm}}{\centering\textbf{Generated Rubric, Distribute}}  \\  % \hline
    \multicolumn{2}{p{13.5cm}}{
    The sample answer is : \textbf{[Sample Answer]}
    
    The grading criteria is :
    
     1. For correctly calculating the average response time and turnaround time for SJF (6.5 points) 
     
     2. For correctly calculating the average response time and turnaround time for FIFO (6.5 points) 
     
     3. For correctly pointing out that if the order of job length is changed, the response time and turnaround time for the SJF policy should remain the same (3 points) 
     
     4. For correctly explaining that the results for the FIFO policy will change with different orders of job length (3 points)
     
     5. For providing a detailed explanation of the execution trace and final statistics for both SJF and FIFO policies with different orders of job length (up to 6 points)
     
     6. For demonstrating an understanding of how SJF and FIFO policies prioritize jobs based on job length and the impact of changing the order of job length on response time and turnaround time (up to 4 points) 
     
     7. For accurately describing the impact of job order on response time and turnaround time for both SJF and FIFO policies (up to 3 points) 
     
     8. For providing a clear and concise explanation of how SJF and FIFO policies handle job execution based on job length and order (up to 2 points) 
     
     9. For demonstrating a basic understanding of the relationship between job arrival order and response time/turnaround time in SJF and FIFO policies (up to 1 point)
    } \\ \hline
\end{tabularx}
\caption{Examples of different rubric types for \textit{Q1} in OS dataset.}
\label{fig:example_rubric}
\end{figure}

%%%%%%% 
\subsection{Evaluation Metric}
\label{Appendix:eval_metric}
The performance of the automated assessment task for short answer questions is evaluated using standard metrics commonly employed in previous ASAG systems. The metrics used to measure grading performance are as follows:
\begin{itemize}
    \item \emph{Mean Absolute Error (MAE)}: \( MAE = \frac{1}{N} \sum_{i=1}^{N} |s_i - \hat{s}_i| \). MAE is known as a scale-dependent accuracy metric and cannot be used to make comparisons between series with different scales.

    \item \emph{Root-Mean-Squared Error (RMSE)}: \(RMSE = \sqrt{\frac{1}{N} \sum_{i=1}^{N} (s_i - \hat{s}_i)^2}\). RMSE is a widely used general-purpose error metric for numerical predictions and provides an absolute error measure.

    \item \emph{Normalized Root-Mean-Squared Error (NRMSE)}: \(NRMSE = \frac{RMSE}{s}\). While RMSE gives an absolute error metric, it is dependent on the absolute scale of the distribution. NRMSE scales the error based on the true mean of the distribution, allowing for performance comparison across different datasets and observing error variations when the dataset changes.
    
    \item \emph{Pearson Correlation Coefficient}: This coefficient evaluates the strength of the linear relationship between the predicted score and the actual score. The coefficient value typically ranges from -1 to +1, but in our experiments, it ranges from 0 to 1 due to the positive relationship between predicted and actual scores. A high Pearson value indicates a strong correlation, demonstrating that the model grades accurately.

    \item \emph{Accuracy}: This metric represents the ratio of correctly identified anomalous student answer-score pairs to the total number of student answer-score pairs.
    
\end{itemize}

%%%%%%%%%%%%%%%%%%%%%%%%%%%%%%%%%%%%%%%%%%%%%%%%%%%%%%%%%%%%
% \newpage
% \input{Styles/checklist}

\end{document}